\pdfoutput=1

\documentclass[11pt]{article}
\usepackage{enumitem}

\usepackage{amsmath}
\usepackage{graphicx}
\usepackage{booktabs}
\usepackage{multirow}
\usepackage{subcaption}
\usepackage{arydshln}
\usepackage{acl}
\usepackage{makecell}
\usepackage{times}
\usepackage{latexsym}

\usepackage[T1]{fontenc}

\usepackage[utf8]{inputenc}

\usepackage{microtype}

\title{HiStruct+: Improving Extractive Text Summarization\\ with Hierarchical Structure Information}

\author{Qian Ruan \\
  DFKI GmbH, Germany \\
  klararuan@gmail.com \\ \And
  Malte Ostendorff  \\
  DFKI GmbH, Germany \\
  malte.ostendorff@dfki.de \\\And
  Georg Rehm \\
  DFKI GmbH, Germany \\
  georg.rehm@dfki.de}

\begin{document}
\maketitle
\begin{abstract}

Transformer-based language models usually treat texts as linear sequences. However, most texts also have an inherent hierarchical structure, i.\,e., parts of a text can be identified using their position in this hierarchy. In addition, section titles usually indicate the common topic of their respective sentences. We propose a novel approach to formulate, extract, encode and inject hierarchical structure information explicitly into an extractive summarization model based on a pre-trained, encoder-only Transformer language model (HiStruct+ model), which improves SOTA
ROUGEs for extractive summarization on PubMed and arXiv substantially. Using various experimental settings on three datasets (i.\,e., CNN/DailyMail, PubMed and arXiv), our HiStruct+ model outperforms a strong baseline collectively, which differs from our model only in that the hierarchical structure information is not injected.  It is also observed
that the more conspicuous hierarchical structure the dataset has, the larger improvements
our method gains. The ablation study demonstrates that the hierarchical position information is the main contributor to our model’s SOTA performance.

\end{abstract}

\section{Introduction}
\label{sec:Introduction}
Texts, especially long documents, contain internal hierarchical structure like sections, paragraphs, sentences, and tokens. When we manually summarize a text, the hierarchical text structure usually plays a key role. Taking a scientific paper as an example, we might focus more on the sections with the titles of  ``methodology'', ``discussion'', and ``conclusion'' while paying less attention to the sections like ``background''. Furthermore, the sentences within one section could have closer relationship with each other, than the ones outside this section. Understanding not only the sequential relations between the sentences but also the internal hierarchical text structure helps us better determine the important sentences within a document. Similarly, a neural summarization model could benefit from these hierarchical structure information.

In this paper, we focus on extractive text summarization of single documents, which is the task of binary sentence classification with labels indicating whether a sentence should be included in a summary. Recently, pre-trained language models based on Transformer \cite{transformer}, such as BERT \cite{bert}, have been widely used to extract contextual representations from texts. The pre-trained Transformer language models (TLMs) can be easily reused for fine-tuning on the downstream tasks, so that the representations already learned from the large pre-training corpora are preserved. \citet{presumm} have achieved the state-of-the-art (SOTA) performance by fine-tuning BERT for extractive summarization on short document datasets including CNN/DailyMail. However, the TLMs consider merely the sequential-context-dependency by adding a linear positional encoding to each input token embeddings. The hierarchical text structure information is not taken into account explicitly.

\begin{figure*}[ht]
  \centering
  \includegraphics[width=1\textwidth]{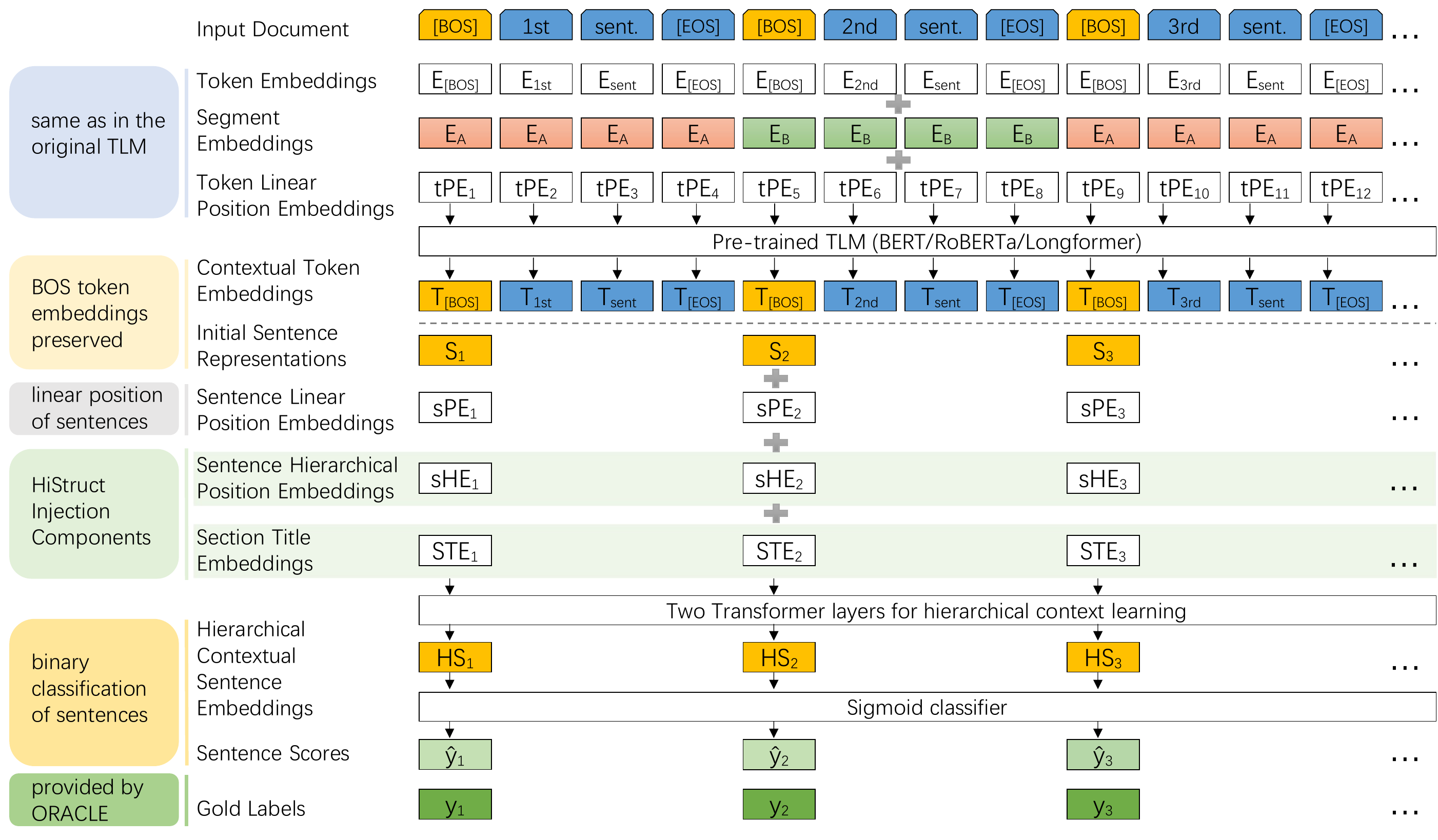}
  \caption[Architecture of the HiStruct+ model]{Architecture of the HiStruct+ model. The  model consists of a base
TLM for sentence encoding and two stacked inter-sentence Transformer layers for hierarchical contextual learning with a sigmoid classifier for extractive summarization. The two blocks shaded in light-green are the HiStruct injection components.}
  \label{fig:overview}
\end{figure*}

We propose a novel approach to formulate, extract, encode and inject the hierarchical structure (HiStruct) information explicitly into an extractive summarization model (HiStruct+ model), which consists of a TLM for sentence encoding and two stacked inter-sentence Transformer layers for hierarchical learning and extractive summarization (see Figure~\ref{fig:overview}). 
We experiment with BERT \cite{bert}, RoBERTa \cite{roberta}, and Longformer \cite{longformer} as underlying TLMs.
The HiStruct information utilized in our work includes the section titles and the hierarchical positions of sentences, which are encoded using our proposed novel methods. 
The resulting embeddings can be injected into the TLM sentence representations to provide the HiStruct information for the summarization task.

The HiStruct+ models are evaluated on short documents (i.\,e., CNN/DailyMail \cite{cnndm}) and long documents (i.\,e., PubMed and arXiv \cite{pubmed}) with various hierarchical characteristics. Our models produce competitive results on CNN/DailyMail and set the SOTA ROUGEs for extractive summarization on PubMed and arXiv to a new level.  
We also compare the HiStruct+ models with the corresponding strong baselines, which differ from our models only in that the HiStruct information is not injected. 
Using various experimental settings, our models collectively outperform the baselines on the three datasets, indicating the effectiveness of the proposed HiStruct encoding methods. The improvements are especially substantial on PubMed and arXiv, which contain longer scientific papers with conspicuous hierarchical structures. Ablation studies suggest that the performance gains are mainly contributed by the hierarchical position information of sentences. %

Our contributions in this work are four-folds: 
(1) We conceptualize novel measures to compare the internal hierarchical structure of the datasets. 
(2) We propose novel methods to formulate the HiStruct information and implement data preprocessing to extract them from the raw datasets.
(3) We propose novel methods to encode and inject the HiStruct information into an extractive summarization model explicitly. The effects of different encoding settings and injection settings are systematically investigated. 
(4) The data containing the extracted HiStruct information,  the best HiStruct+ models, as well as the scripts for preprocessing, training and evaluation are available on GitHub\footnote{\url{https://github.com/QianRuan/histruct}}.  %

\section{Related Work}
\label{sec:Related Work and background}

\subsection{Text Summarization}
\label{subsec:text_summarization}

\textbf{Extractive Text Summarization} (ETS) is to 
classify sentences within a document with labels indicating whether a sentence should be included in the summary. 
\citet{presumm} fine-tune BERT with stacked Transformer layers and a sigmoid classifier (BERTSUMEXT). Instead of directly utilizing the existing Transformer encoder for document encoding, \citet{zhang-etal-2019-hibert} pre-train a hierarchical Transformer encoder consisting of a sentence encoder and a document encoder (HIBERT) and fine-tune it for ETS. 
For long documents, \citet{xiao-carenini-2019-extractive} propose a RNN-based ETS model incorporating both the global and the local context (ExtSum-LG). 
To address the problem of redundancy in extractive summaries, the authors further improve their work by introducing redundancy reduction  \cite{xiao-carenini-2020-systematically}. They systematically explore and compare different methods including Trigram Blocking \cite{paulus2018a}, RdLoss, MMR-Select and MMR-Select+ \cite{xiao-carenini-2020-systematically}. Trigram Blocking is a traditional redundancy reduction method that avoids adding a candidate sentence to the summary if it has trigram overlap with the previously selected sentences.
Their previous model combined with the redundancy reduction methods produce SOTA performance for ETS on PubMed and arXiv \cite{xiao-carenini-2020-systematically}.

Previous works on extractive summarization model hierarchical structure of documents by introducing a hierarchical attention, where they first learn contextual token representations based on the linear dependencies between tokens and then add additional CNN  \cite{cheng-lapata-2016-neural} or RNN \cite{summarunner}  or Transformer \cite{zhang-etal-2019-hibert, presumm} layer(s) to learn document-level representations for each sentence based on the linear dependencies between sentences. However, they learn hierarchical representations of sentences in an implicit way. The models are like black boxes, lacking interpretability. In contrast, our proposed approach enriches sentence representations in an explicit way by using section titles and hierarchical positions of sentences as additional HiStruct information, which is more intuitive and interpretable. 

\textbf{Abstractive text summarization} (ATS) is to generate summaries with new sentences which are not present in the source text. 
BERTSUMABS \cite{presumm} uses the pre-trained BERT as the encoder in its encoder-decoder architecture. Instead of simply using the pre-trained BERT, recent works, including T5 \cite{t5}, BART \cite{bart} and PEGAUSUS \cite{zhang2019pegasus} pre-train encoder-decoder models specifically for seq2seq tasks. The first attempt at addressing neural abstractive summarization of long documents is undertaken by \citet{pubmed}. %
\citet{aksenov2020i} overcome the length limitations of BERT by a new method of BERT-windowing, allowing it to deal with longer documents.
\citet{dancer2020} propose a divide-and-conquer approach to train a model to summarize each part of the document separately. To address the essential issue of the quadratic full attention operation of TLMs, \citet{bigbird} propose BigBird with a sparse attention mechanism.

\textbf{Hybrid text summarization} combines extractive summarization, abstractive summarization, or other techniques as a two-stage hybrid system.
 MatchSum \cite{zhong-etal-2020-extractive} is a recent work that first selects sentences from a document using an extractive model and builds a set of candidate summaries based on them. The summarization task is then formulated as a semantic text matching problem between the source document and the candidate summaries.  \citet{pilault-etal-2020-extractive} presents a hybrid system that consists of an extractive model and a Transformer language model. The Transformer language model employs an encoder-decoder architecture for abstractive summarization, conditioned on the sentences extracted by the extractive model.

\subsection{Injection of Additional Information}
\label{subsec:Hierarchical learning}

The idea of injecting additional information to TLM is inspired by two former works, LAMBERT \cite{lambert} and LayoutLM \cite{layoutlm}, where the visual layout information is injected into BERT by adjusting its input embeddings. 
These models were not proposed for text summarization and they cannot be applied to plain texts since the layout positions have to be obtained from scanned document images. In contrast, our approach makes use of the internal HiStruct information, which can be found in most types of textual data. Moreover, we enrich the output representations from the TLM instead of adjusting the input embeddings. This saves compute resources since TLM pre-training is not required.

\section{Methodology}
\label{sec:Methodology}

\subsection{Hierarchical Structure Information}
\label{subsec:hierarchical_structure_features}

\textbf{Hierarchical position} of a sentence is represented in the proposed method as a vector of its positions at each hierarchy-level. 
\begin{equation} \label{eq:sentstructvec}
SSV_s = (a_s, b_s)
\end{equation}
Given the $s$-th sentence within a document, its hierarchical position is formulated as a 2-dimensional vector $(a_s,b_s)$, denoted as the sentence structure vector $SSV_{s}$, where $a_s$ represents the linear position of the section containing the sentence and $b_s$ is the linear position of the sentence within the section. All sentences within the same section have the same value in the first dimension of the SSV, indicating the close relationships between them. The second dimension indicates more precisely their linear relations within the section. By this very simple numerical formulation, hierarchical relations between sentences are clearly identified.

\textbf{Section titles} exist in particular in long documents like scientific papers. They usually imply the section content and describe the common topic for its sub-sentences \cite{ostendorff-etal-2020-aspect}. 
In our work, we propose to utilize the corresponding section title as an additional HiStruct information when encoding its sub-sentences. There exist typical section titles in scientific papers. Similar section titles like ``Conclusion'', ``Conclusions'' and ``Concluding remarks''  have the same semantic meaning and can be grouped into one typical section title class of ``Conclusions''. This is also taken into consideration when encoding the section titles. 

\subsection{Hierarchical Structure Encoding}
\label{subsec:hierarchical_structure_encoding}

\textbf{Hierarchical position embedding} is based on the existing linear position encoding methods (PE), including the sinusoidal method (sin) used by Transformer \cite{transformer} and the learnable method (la) used by BERT \cite{bert}. We use one of the PEs to encode the two dimensions of a SSV respectively, resulting in two embeddings. Using the la PE, the embeddings are initialized randomly and trained with the entire summarization model. Using the sin PE, the two embeddings are calculated simply by Equations \ref{eq:sinusoid1} and \ref{eq:sinusoid2} as described by \citet{transformer}. 

\begin{equation}\label{eq:sinusoid1}
\footnotesize
PE_{(pos, 2i)} =sin(pos/10000^{2i/d_{model}})
\end{equation}
\begin{equation}\label{eq:sinusoid2}
\footnotesize
PE_{(pos, 2i+1)} =cos(pos/10000^{2i/d_{model}})
\end{equation}
\noindent
where $pos$ is the value in one dimension of the SSV and $i$ is the $i$-th dimension of the resulting embedding.

Given the $s$-th sentence with the hierarchical position of $(a_s,b_s)$, and the desired size of the output embeddings $d$, the Sentence Hierarchical Position
Embedding (sHE) can be generated by Equations \ref{eq:sHE_sum}, \ref{eq:sHE_mean}, \ref{eq:sHE_concat}, using different combination modes. 
\begin{equation} 
\label{eq:sHE_sum}
\footnotesize
sHE_{\text{sum}} (s, d) = PE(a_s, d) \\
+ PE(b_s, d)
\end{equation}
\begin{equation}
\footnotesize
\label{eq:sHE_mean}
sHE_{\text{mean}} (s, d) = \frac{PE(a_s, d) + PE(b_s, d)}{2}
\end{equation}
\begin{equation}
\footnotesize
\label{eq:sHE_concat}
sHE_{\text{concat}} (s, d) = PE(a_s,\frac{d}{2})  |  PE(b_s,\frac{d}{2}) 
\end{equation}
where the symbol | denotes vector concatenation.

Using one of the PEs (i.\,e., sin or la) associated with one of the combination modes (i.\,e., sum, mean or concat), it totals six different settings of the hierarchical position encoding method: sin-sum, sin-mean, sin-concat, la-sum, la-mean and la-concat. %

\textbf{(Classified) section title embedding} is generated by the same pre-trained TLM, which is involved in the summarization model. We have two options to encode section titles: section title embedding (STE) and classified section title embedding (classified STE). 
A section title embedding is generated by feeding the tokenized section title into the TLM and summing up the last hidden states at each token position as a single embedding. Similar section titles consisting of similar tokens lead to embeddings that are already similar to each other in some way. We also manually pre-define typical section title classes and the corresponding intra-class section titles depending on the datasets and the domains. Using the classified STE, all intra-class STEs are replaced with the embedding of its corresponding class. In the case that a section title does not belong to any class or it falls into more than one class, the original STE is used.  %

\subsection{Model Architecture}
\label{subsec:Model architecture}

Figure \ref{fig:overview} illustrates the overview architecture of the proposed HiStruct+ model. The model consists of a base TLM for sentence encoding and two stacked inter-sentence Transformer layers for hierarchical learning and extractive summarization. The sequence on top is the input document, tokenized by the corresponding tokenizer of the involved TLM. The input embeddings to the TLM are the same as in the original TLM. In order to represent individual sentences, we insert a BOS token at the start of every sentence. 
Only the BOS token embeddings are preserved as the initial sentence representations ($S_s$). Each sentence representation is first enriched with a Sentence Linear Position Embedding, which encodes its linear position within the whole document. An additional Sentence Hierarchical Position Embedding ($sHE_s$) can be added, which is generated by encoding the hierarchical position of the sentence using the proposed hierarchical position encoding method. If section titles are available, we can further enrich the sentence representation by adding a STE or classified STE ($STE_s$). The sentence representations with the injected HiStruct information are fed to the two stacked Transformer encoder layers to learn inter-sentence document-level hierarchical contextual features. 
The result is a set of Hierarchical Contextual Sentence Embeddings ($HS_s$). The final output layer is a sigmoid classifier, which calculates the confidence score $\hat{y}_s$ of including the $s$-th sentence in the extractive summary based on the $HS_s$. The loss of the summarization model is the binary classification entropy of the prediction $\hat{y}_s$ against the gold label $y_s$.

The two HiStruct injection components shaded in light-green are optional. Removing these from the HiStruct+ model based on BERT, the architecture is identical to BERTSUMEXT \cite{presumm}, which is a strong baseline against our models on CNN/DailyMail. When using RoBERTa and Longformer as the base TLM, we also construct a baseline model without the two components. The comparison baselines are named as TransformerETS in this paper. The effectiveness of injecting HiStruct information using the proposed methods can be systematically investigated by comparing our HiStruct+ model to the corresponding TransformerETS baseline which uses the same base TLM and the same input length, but is unaware of the HiStruct information.

\section{Experimental Setup}
\label{sec:Experiments}
\subsection{Datasets}
\label{subsec:datasets}

Our models are evaluated on three benchmark datasets for single document summarization, including CNN/DailyMail \cite{cnndm}, PubMed and arXiv \cite{pubmed}. Table \ref{tab:datasets} presents detailed statistics of the datasets.

The three datasets represent different document types ranging from short news articles to long scientific papers. To emphasize the difference in the hierarchical structure among different datasets, we define the concepts of hierarchical depth (hi-depth) and hierarchical width (hi-width). The hi-depth refers to the number of the hierarchy-levels within the document. Scientific papers have a deeper hierarchy consisting of sections, paragraphs, sentences and tokens (i.\,e., hi-depth = 4). In news articles, paragraphs are not further grouped into sections (i.\,e., hi-depth = 3). In this case, we use paragraphs instead of sections as the highest hierarchy level when representing the hierarchical position of sentences (i.\,e.,  the first dimension of the SSVs).
The hierarchical width, $\textnormal{hi-width} = {\dfrac{N_s} {N_{hsh}}}$, is the ratio of total number of sentences $N_s$ and the number of the text-units regarding the highest structure hierarchy $N_{hsh}$. 
It indicates how many sentences are there on average in every paragraph/section. The more sentences are there, the second dimension of the SSVs has a more wide range of values, and the values in the first dimension of the SSVs differ a lot from the linear sentence positions. Larger hi-depth and larger hi-width indicate that the hierarchical structure of the dataset is more conspicuous.

We hypothesize that the proposed method works better on datasets with more conspicuous hierarchical structures, where hi-depth and hi-width are larger. This will be proved by comparing the performance improvements on the three datasets with different hierarchical characteristics. %

\textbf{CNN/DailyMail} is included as an exemplary dataset with less conspicuous hierarchical structure compared to PubMed and arXiv. The average hi-width over all documents is 1.33, which is much smaller than those in PubMed and arXiv. The dataset contains more than 310k news articles. We use the standard splits given by \citet{cnndm} for training, validation, and testing. 

During data preprocessing, we first split documents into sentences and paragraphs respectively with the Stanford CoreNLP toolkit \cite{manning-etal-2014-stanford}. The sentences and paragraphs are tokenized, resulting in the lists of sentence tokens and the lists of paragraph tokens. SSVs corresponding to each sentence can be obtained by comparing those lists side by side. For all three datasets, we use a greedy selection algorithm similar to \citet{summarunner} and \citet{presumm} to select sentences from documents as the gold extractive summaries (ORACLE). Sentences in the ORACLE summaries are assigned with the gold label 1. 

\textbf{PubMed and arXiv} contain longer scientific papers. PubMed contains papers in the bio-medical domain, while arXiv contains papers in various domains.  The average hi-width over all PubMed documents is 15.79, in arXiv it is 37.33. 
We use the original splits  given by \citet{pubmed} for training, validation, and testing. SSVs are obtained by tokenizing the sentences and sections of every document respectively. The details on the generation of section title embeddings and classified section title embeddings can be found in Appendix \ref{subsec:Pre-defined section classes}.

\subsection{Implementation Details}
\label{subsec:details}
We implement our model based on BERTSUMEXT \cite{presumm}
and use HuggingFace Transformers \cite{wolf-etal-2020-transformers} to make use of the pre-trained instances of BERT, RoBERTa and Longformer. 
On CNN/DailyMail, we select 3 sentences with Trigram Blocking. On PubMed and arXiv, 7 sentences are extracted while Trigram Blocking is not applied (see more details with regard to implementation in Appendix~\ref{subsec:Implementation Details appendix} and~\ref{subsec:Experimental settings}).

\section{Results and Discussion}
\label{sec:Results and Discussion}
We evaluate the performance of our summarization models automatically using ROUGE metrics \cite{rouge} including F1 ROUGE-1 (R1), ROUGE-2 (R2) and ROUGE-L (RL). Tables \ref{tab:cnndm_result}, \ref{tab:pubmed_result} and \ref{tab:arxiv_result} summarize the performance of our models in comparison to the baselines and the previously reported SOTA results on CNN/DailyMail, PubMed and arXiv respectively. On all three datasets, ablation studies are systematically conducted to investigate the contributions of different experimental settings. To analyze the output summaries from an overall perspective, we plot the distribution of the extracted sentences on each dataset and compare it to the ORACLE summaries and those outputted by the comparison baseline (see Figure \ref{fig:extracted summaries}). Appendix \ref{subsec:human_evluation} demonstrates human evaluation of extracted summaries for a more intuitive understanding about the superiority of the proposed system.

\subsection{Results on CNN/DailyMail}
\label{subsec:Results on short documents}

\begin{table}[ht]
\fontsize{9}{9}
\selectfont
\centering
        \begin{tabular}[t]{@{}llll@{}}
        \toprule
         Model $\downarrow$ / Metric $\rightarrow$  & \multicolumn{1}{c}{R1} & \multicolumn{1}{c}{R2} & \multicolumn{1}{c}{RL} \\ \midrule
        \multicolumn{4}{c}{Abstractive}                 \\ \midrule
BERTSUMABS \citeyearpar{presumm}              & 41.72 & 19.39 & 38.76 \\
BART \citeyearpar{bart}                    & 44.16 & 21.28 & 40.90 \\
PEGASUS \citeyearpar{zhang2019pegasus}                 & \underline{44.17} & \underline{21.47} & \underline{41.11} \\
BigBird PEGASUS \citeyearpar{bigbird}         & 43.84 & 21.11 & 40.74 \\ \midrule
        \multicolumn{4}{c}{Extractive}                  \\ \midrule
HIBERT \citeyearpar{zhang-etal-2019-hibert}    &  &  & \\
~ (BERT-base)  & 42.31 & 19.87 & 38.78 \\
~ (BERT-large) & 42.37 & 19.95 & 38.83 \\

BERTSUMEXT \citeyearpar{presumm}  &  &  & \\
~ (BERT-base)  & 43.25 & 20.24 & 39.63 \\
~ (BERT-large) & \underline{43.85} & \underline{20.34} & \underline{39.90}  \\ 
\midrule
        \multicolumn{4}{c}{Hybrid}                      \\ \midrule
MatchSum \citeyearpar{zhong-etal-2020-extractive} &  &  & \\
~ (BERT-base)  & 44.22 & 20.62 & 40.38 \\
~ (RoBERTa-base) & \underline{44.41} & \underline{20.86} & \underline{40.55} \\ \midrule
\multicolumn{4}{c}{Reproduced baselines}                                                       \\ \midrule
ORACLE (512 tok.)              & 52.46  & 30.76  & 48.66  \\
ORACLE (1,024 tok.)               & 55.45  & 32.78  & 51.59  \\
LEAD-3                                & 40.33  & 17.39  & 36.56  \\
TransformerETS  & & & \\
~~ \textit{BERT-base (1,024 tok.)}          & 43.32  & 20.27  & 39.69    \\
~~ \textit{BERT-large (512 tok.) }           & 43.45  & 20.36  & 39.83    \\
~~ \textit{RoBERTa-base (1,024 tok.)}        & 43.62  & 20.53  & 39.99    \\ \midrule
\multicolumn{4}{c}{Our models (Extractive)}                                          \\ \midrule
HiStruct+  & & & \\
~~ \textit{BERT-base (1,024 tok.)} & \textbf{43.38}  & \textbf{20.33}  & \textbf{39.78}    \\
~~ \textit{BERT-large (512 tok.)}  & \textbf{43.49}  & \textbf{20.40}*  & \textbf{39.90}*  \\
~~ \textit{RoBERTa-base (1,024 tok.)}   & \underline{\textbf{43.65}} & \underline{\textbf{20.54*}} & \underline{\textbf{40.03*}}  \\ \midrule
\multicolumn{4}{c}{Our models (Hybrid)} \\ \midrule
HiStruct+ \\\textit{RoBERTa-base (1,024 tok.)}
\\ \& MatchSum (RoBERTa-base)
& 44.31 & 20.73  & 40.47  \\ \bottomrule
\end{tabular}
\caption[Results on CNN/DailyMail]{F1 ROUGE results on CNN/DailyMail. Bold are the scores of the HiStruct+ models that are better than the corresponding TransformerETS baseline. The symbol * indicates an improvement over the corresponding SOTA ROUGE for extractive summarization. }
\label{tab:cnndm_result}
\end{table}

\begin{figure}[ht]
\begin{subfigure}{1\linewidth}
  \centering
  \includegraphics[width=1\textwidth]{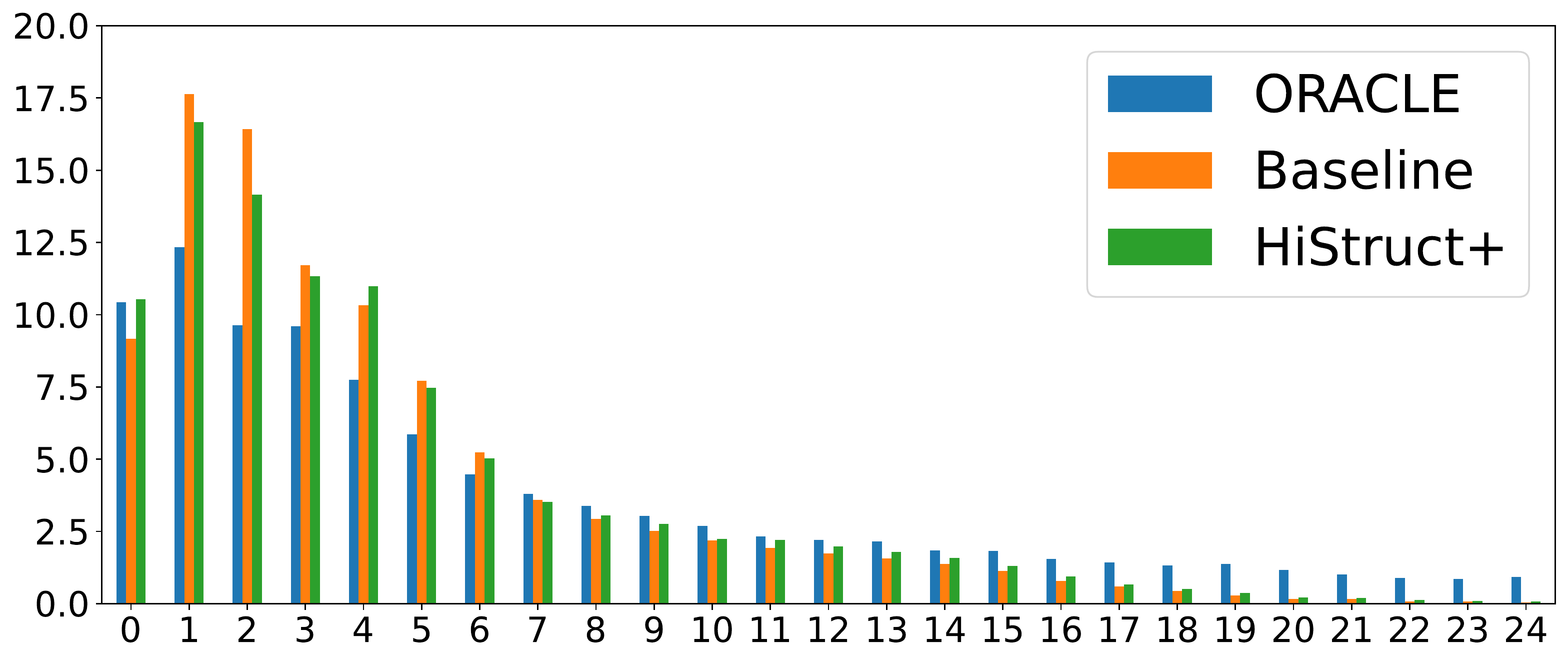}
  \caption[CNN/DailyMail: distribution of the extracted sentences]{CNN/DailyMail}%
  \label{fig:cnndm_dist_roberta1024}
\end{subfigure}
\begin{subfigure}{1\linewidth}
  \centering
  \includegraphics[width=1\textwidth]{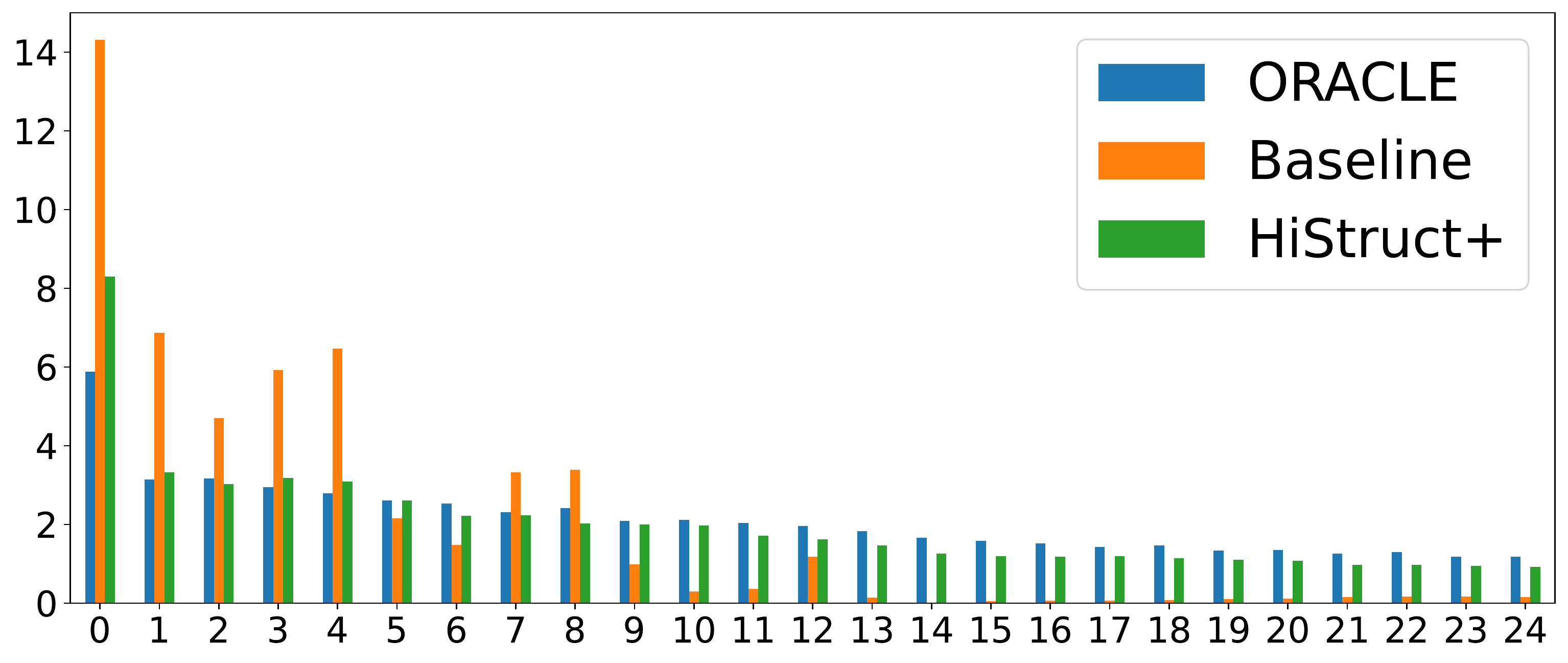}
  \caption{PubMed}%
  \label{fig:pubmed_dist_longformer15000}
\end{subfigure}
\begin{subfigure}{1\linewidth}
  \centering
  \includegraphics[width=1\linewidth]{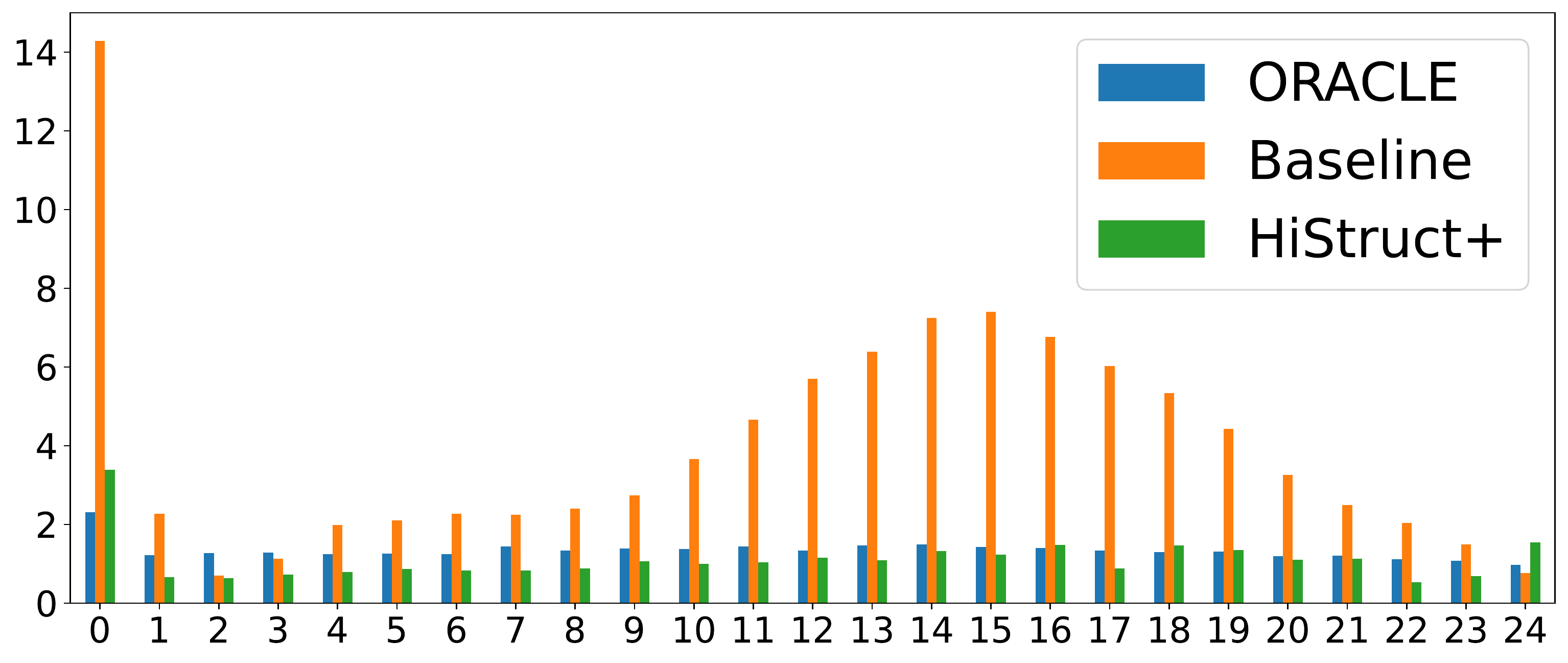}
  \caption[arXiv: distribution of the extracted sentences]{arXiv}%
  \label{fig:arxiv_dist_longformer28000}
\end{subfigure}
\caption{Proportions of the extracted sentences at each linear position. The x-axis values are linear sentence indices, the y-axis values are percentages of the extracted sentences. In this figure, only the first 25 sentence indices are included due to space limitation.}
\label{fig:extracted summaries}
\end{figure}

\textbf{ROUGE results} on CNN/DailyMail are summarized in Table \ref{tab:cnndm_result}. 
The first three blocks highlight the results reported by the corresponding papers of abstractive, extractive, and hybrid summarization systems. The best results regarding the respective type of the summarization system are underlined. 
In the baselines block, the first two lines highlight the ORACLE results that build the upper bounds for extractive systems taking the same number of input tokens. %
The LEAD-n baselines simply select the first n sentences in a document as its extractive summary. Despite its simplicity, the LEAD-3 baseline already achieves relatively competitive performance.
The three TransformerETS models are the corresponding comparison baselines that use the same model architecture and experimental settings as our models but without injected HiStruct information.
The following block presents the results of our HiStruct+ models based on different TLMs with various input lengths. To make the evaluation results comparable to the SOTA extractive model BERTSUMEXT, we follow their approach and report the averaged results of three best checkpoints. 

Regardless the base TLM and input length, our HiStruct+ models collectively outperform the corresponding TransformerETS baselines by merely injecting the hierarchical position information of sentences. 
However, the performance improvements gained by our models on CNN/DailyMail are small. One of the reasons might be that we merely inject the hierarchical position information of sentences, section titles are not available. Furthermore, as discussed in Section \ref{sec:Experiments}, the hierarchical structure of the CNN/DailyMail documents is not so obvious as those in PubMed and arXiv. %

Compared to the SOTA extractive model, our best HiStruct+ model produces competitive results. The R2 and RL scores are improved slightly. 
The model can be reused in many hybrid approaches. When we apply MatchSum based on our best model, the ROUGE results are further increased.

\textbf{Ablation studies} on CNN/DailyMail (see the  results and detailed discussions in Appendix \ref{subsec:Ablation studies on CNN/DailyMail}) suggest that the setting la-sum works best for hierarchical position encoding. Two stacked Transformer layers in the summarization model perform better than one or three Transformer layers. When taking longer inputs than the length limit of the TLM, substantial improvements are achieved by using the copied token position embeddings for initialization instead of random initialization. 
\\
\indent
\textbf{The extracted summaries} are analyzed in more detail by plotting the proportions of the extracted sentences at each linear position within the whole document as shown in Figure \ref{fig:cnndm_dist_roberta1024}. The model in green is our best-performed HiStruct+ model on CNN/DailyMail. The model in orange is the corresponding comparison baseline without injected HiStruct information. The model in blue is the ORACLE system, which produces the gold extractive summaries. We can observe that the ORACLE summary sentences are distributed across documents more smoothly, while our HiStruct+ model and the baseline model tend to select the first sentences and fail to select sentences that appear at later positions within the documents. Compared to the baseline, the HiStruct+ model leads to more similar proportions as the ORACLE summaries at the most sentence indices.

\subsection{Results on PubMed}
\label{subsec:Results on PubMed}

\begin{table}[ht]
\fontsize{9}{9}
\selectfont
\centering
\begin{tabular}[t]{@{}llll@{}}
\toprule
Model $\downarrow$ / Metric $\rightarrow$  & \multicolumn{1}{c}{R1} & \multicolumn{1}{c}{R2} & \multicolumn{1}{c}{RL} \\ \midrule
\multicolumn{4}{c}{Abstractive}               \\ \midrule
PEGASUS \citeyearpar{zhang2019pegasus}   & 45.49 & 19.90  & \underline{42.42} \\
BigBird PEGASUS \citeyearpar{bigbird}   & 46.32 & \underline{20.65} & 42.33 \\
DANCER PEGASUS \citeyearpar{dancer2020}   & \underline{46.34} & 19.97 & \underline{42.42} \\ \midrule
\multicolumn{4}{c}{Extractive}                \\ \midrule
Sent-CLF \citeyearpar{pilault-etal-2020-extractive}              & 45.01 & 19.91 & \underline{41.16} \\
Sent-PTR \citeyearpar{pilault-etal-2020-extractive}              & 43.30  & 17.92 & 39.47 \\
ExtSum-LG+ \citeyearpar{xiao-carenini-2020-systematically}   & & &\\
~ RLoss      & 45.30  & \underline{20.42} & 40.95 \\
~ MMR-Select+ & \underline{45.39} & 20.37 & 40.99 \\ \midrule
\multicolumn{4}{c}{Hybrid}                    \\ \midrule
TLM-I+E(G,M) \citeyearpar{pilault-etal-2020-extractive}       & \underline{42.13} & \underline{16.27} & \underline{39.21} \\ \midrule %
\multicolumn{4}{c}{Reproduced baselines}                           \\ \midrule
ORACLE (4,096 tok.)     & 49.73   & 27.29 & 45.26  \\
ORACLE (9,600 tok.)      & 52.80    & 28.95 & 48.08  \\
ORACLE (15k tok.)      & 53.04   & 29.08 & 48.31  \\
LEAD-7                       & 38.30    & 12.54 & 34.31  \\
LEAD-10                      & 38.59   & 13.05 & 34.81  \\
TransformerETS   & & & \\
~~ \textit{Longformer-base (15k tok.)}   & 41.69   & 15.76 & 37.48  \\
~~ \textit{Longformer-large (15k tok.)}  & 41.69   & 15.79 & 37.49  \\ \midrule
\multicolumn{4}{c}{Our models (Extractive)}   \\ \midrule
HiStruct+  &&&\\
~~ \textit{Longformer-base (15k tok.)}  &&&  \\ 
~~~~~~ sHE+STE(classified) & \underline{\textbf{46.59*’}} & \underline{\textbf{20.39}} & \underline{\textbf{42.11*}} \\
~~~~~~ sHE+STE                      & \textbf{46.49*’} & \textbf{20.29} & \textbf{42.02*} \\
~~~~~~ sHE                          & \textbf{45.76*}  & \textbf{19.64} & \textbf{41.34*} \\ %
~~ \textit{Longformer-large (15k tok.)} &&& \\ %
~~~~~~ sHE+STE(classified)          & \textbf{46.38*’} & \textbf{20.17} & \textbf{41.92*} \\
~~~~~~ sHE                          & \textbf{45.67*}  & \textbf{19.60}  & \textbf{41.26*} \\ 
\bottomrule
\end{tabular}
\caption[Results on PubMed]{F1 ROUGE results on PubMed. Bold are the scores of the HiStruct+ models that are better than the corresponding TransformerETS baseline. The symbol * indicates that the corresponding SOTA ROUGE for extractive summarization is improved by our model. The symbol ' indicates that the SOTA ROUGEs (incl. all types of summarization approaches) are outperformed.}
\label{tab:pubmed_result}
\end{table}

\textbf{ROUGE results} on PubMed are summarized in Table \ref{tab:pubmed_result}. 
As shown in the baselines block, the ORACLE upper bounds for extractive summarization are increased significantly by increasing the input length, which makes it possible to exploit potential gains from modeling longer input. The LEAD-n baselines do not produce competitive results on PubMed. It indicates that the first sentences in PubMed are not so informative as those in CNN/DailyMail. The last two TransformerETS models in the block are the comparison baselines that are unaware of HiStruct.

The last block in Table \ref{tab:pubmed_result} presents the results of two groups of HiStruct+ models, grouped by the base TLM used in the summarization model. In PubMed, we can choose to inject the sentence hierarchical position embeddings (sHEs) with or without the section title embeddings (STEs). STEs can be replaced by classified STEs. This can result in three different injection settings for a model group, namely sHE, sHE+STE, and sHE+STE(classified). For each model setting, we report the results of the best-performed checkpoint. 

Our best HiStruct+ model on PubMed is a model based on Longformer-base taking 15,000 input tokens, which injects the sHEs and the classified STEs into the extractive model. It achieves ROUGE results of 46.59/20.39/42.11, which beat the SOTA extractive model ExtSum-LG+MMR-Select+ collectively on all three ROUGE metrics with improvements of 1.2/0.02/1.12. Taking the SOTA abstractive and hybrid approaches into account, our results are still very competitive.

All HiStruct+ models produce the competitive results that are better than or very close to the former SOTA results for extractive summarization. They also collectively outperform the TransformerETS baselines by a large margin on all evaluation metrics. The overperformance is much more substantial than that on CNN/DailyMail, even if only the hierarchical position information is injected. 
This supports our hypothesis that the proposed model works better on datasets with more conspicuous hierarchical
structures.

\textbf{Ablation studies} on PubMed suggest that the largest improvement of our models against the baseline is contributed by the hierarchical position information of sentences. This is observed when we compare the three models in the first group of HiStruct+ models with the first TransformerETS baseline. Injecting merely sHE, the results are already increased by 4.07/3.88/3.86. When the section title embedding (STE) is included additionally, the results are further increased by 0.73/0.65/0.68. When using classified STE instead, the ROUGEs are increased by a small margin of 0.1/0.1/0.09. Comparing the second group of HiStruct+ models to the second TransformerETS baseline, it is also observed that injecting the sHE leads to the largest performance gain.

\textbf{The extracted summaries} analysis on PubMed test set is demonstrated in Figure \ref{fig:pubmed_dist_longformer15000}. The model in green is our best-performed HiStruct+ model on PubMed, the model in orange is the corresponding TransformerETS baseline, the model in blue is the ORACLE system. The ORACLE summaries are distributed across documents evenly. The TransformerETS baseline favors the first 5 sentences and ignores the sentences appearing at later positions. In contrast, our HiStruct+ model overcomes the problem of focusing merely on the first sentences. The outputs of the HiStruct+ model are close to the ORACLE summaries. It indicates that by injecting HiStruct information explicitly using our method, the model successfully learns the deeper internal hierarchical structure of the PubMed documents and relies less on the linear sentence positions.

\subsection{Results on arXiv}
\label{subsec:Results on arXiv}

\begin{table}[ht]
\fontsize{9}{9}
\selectfont
\centering
\begin{tabular}[t]{@{}llll@{}}
\toprule
Model $\downarrow$ / Metric $\rightarrow$ & \multicolumn{1}{c}{R1} & \multicolumn{1}{c}{R2} & \multicolumn{1}{c}{RL} \\ \midrule
\multicolumn{4}{c}{Abstractive }               \\ \midrule
PEGASUS \citeyearpar{zhang2019pegasus}                & 44.70  & 17.27 & 25.80  \\
BigBird PEGASUS \citeyearpar{bigbird}        & \underline{46.63} & 19.02 & \underline{41.77} \\
DANCER  PEGASUS \citeyearpar{dancer2020}       & 45.01 & 17.60  & 40.56 \\
LED-large %
\citeyearpar{longformer}    & \underline{46.63} & \underline{19.62} & 41.48 \\ \midrule
\multicolumn{4}{c}{Extractive }                \\ \midrule
Sent-CLF \citeyearpar{pilault-etal-2020-extractive}              & 34.01 & 8.71  & 30.41 \\
Sent-PTR \citeyearpar{pilault-etal-2020-extractive}              & 42.32 & 15.63 & 38.06 \\
ExtSum-LG + \citeyearpar{xiao-carenini-2020-systematically}  &&&\\
~ RLoss & \underline{44.01} & \underline{17.79} & \underline{39.09} \\
~ MMR-Select+  & 43.87 & 17.50  & 38.97 \\ \midrule
\multicolumn{4}{c}{Hybrid }                    \\ \midrule
TLM-I+E(G,M) \citeyearpar{pilault-etal-2020-extractive}       & \underline{41.62} & \underline{14.69} & \underline{38.03} \\\midrule
\multicolumn{4}{c}{Reproduced baselines}                              \\ \midrule
ORACLE (15k tok.)   & 53.58    & 26.19   & 47.76   \\
ORACLE (28k tok.)    & 53.97    & 26.42   & 48.12   \\
LEAD-10                     & 37.37    & 10.85   & 33.17   \\
TransformerETS   & & & \\
~~ \textit{Longformer-base (15k tok.)}   & 38.49    & 11.59   & 33.85   \\
~~ \textit{Longformer-base (28k tok.)}   & 38.47    & 11.56   & 33.82   \\ \midrule
\multicolumn{4}{c}{Our models (Extractive)}      \\ \midrule
HiStruct+  &&&\\
~~ \textit{Longformer-base (15k tok.)}  &&&  \\ 
~~~~~~ sHE+STE(classified)         & \textbf{44.94*}   & \textbf{17.42}   & \textbf{39.90*}   \\
~~~~~~ sHE+STE                     & \textbf{45.02*}   & \textbf{17.48}   & \textbf{39.94*}  \\
~~~~~~ sHE                         & \textbf{43.04}    & \textbf{15.87}   & \textbf{38.13}   \\ 
~~ \textit{Longformer-base (28k tok.)}  &&&  \\ 
~~~~~~ sHE+STE(classified)         & \textbf{45.17*}   & \textbf{17.61}   & \textbf{40.10*}   \\
~~~~~~ sHE+STE                     & \underline{\textbf{45.22*}}   & \underline{\textbf{17.67}}   & \underline{\textbf{40.16*}}  \\ \bottomrule
\end{tabular}
\caption[Results on arXiv]{F1 ROUGE results on arXiv. Bold are the scores of the HiStruct+ models that are better than the corresponding TransformerETS baseline. The symbol * indicates that the corresponding SOTA ROUGE for extractive summarization is improved by our model. }
\label{tab:arxiv_result}
\end{table}

\textbf{ROUGE results} on arXiv are summarized in Table \ref{tab:arxiv_result}. 
The results of the HiStruct+ models are presented in two groups. The first group takes 15k input tokens, while the second group increases the input length to 28k. In the groups, different injection settings are compared. 

Our best-performed HiStruct+ model on arXiv is an extractive model based on Longformer-base with 28k input tokens, injecting the sHEs with the original STEs. This model beats the results achieved by ExtSum-LG+RLoss and sets the new SOTA ROUGEs for extractive summarization on arXiv to 45.22/17.67/40.16. 

All HiStruct+ models collectively outperform the corresponding TransformerETS baselines (i.\,e., the last two models in the baselines block) by a large margin on all ROUGE scores. On this dataset, the HiStruct+ improvement is much more significant than those on both CNN/DailyMail and PubMed. The arXiv dataset has the largest hi-width among the three datasets and the hierarchical structure is most conspicuous, which might be the reason why the HiStruct+ models yield the largest performance improvements on arXiv.

\textbf{Ablation studies} in the first HiStruct+ group also suggest that the largest improvement of our HiStruct+ model against the TransformerETS baseline is contributed by injecting the sentence hierarchical position information, which is encoded as sHEs. %
The effect of using the classified STE on arXiv is opposite to that on PubMed. The summarization performance declines slightly when we replace the STE with the classified STE. This outcome occurs in the second group of HiStruct+ models as well. We notice the fact that there are 500k unique section titles in arXiv, while PubMed contains 164k unique section titles. Accordingly, it becomes much more difficult to group a large number of section titles correctly into several section classes. Furthermore, the PubMed dataset contains papers mostly in the bio-medical domain. The structure of those papers tends to follow specific writing conventions in the bio-medical sciences. The arXiv dataset, in contrast, contains scientific papers that are not limited to a specific domain. As consequence, the document structure and the writing styles are more diverse.

\textbf{The extracted summaries} analysis on arXiv is demonstrated in Figure \ref{fig:arxiv_dist_longformer28000}. 
The baseline (in orange) tends to select the first sentence and the sentences indexed between 10 and 20, while it excludes sentences at later positions. It is clearly observed that the summary sentences extracted by the HiStruct+ model are evenly distributed, the informative sentences appearing at later positions are not ignored.

\section{Conclusions}
\label{sec:Conclusions}
This work addresses hierarchical modeling for extractive text summarization by explicitly leveraging hierarchical structure information, including section titles, as well as hierarchical position information of the sentences.
We propose an intuitive and interpretable approach to formulate, extract, encode and inject the hierarchical structure information into an extractive summarization model.  

The proposed HiStruct+ models are systematically evaluated on CNN/DailyMail, PubMed, and arXiv. 
On PubMed, our model increases the former SOTA ROUGEs for extractive summarization by 1.2/0.02/1.12. On arXiv, the new SOTA results for extractive summarization are set to 45.22/17.67/40.16. Our ablation studies suggest that the SOTA performance are mostly gained by providing the hierarchical position information of sentences to the summarization model. 
When comparing the HiStruct+ models with the baselines that are unaware of the HiStruct information, improvements are consistently observed on all three datasets under various experimental settings, indicating the effectiveness of the proposed method. Moreover, our experiments show that the more conspicuous hierarchical structure the dataset has, the larger the improvements of our method are.
The proposed metrics of hi-depth and hi-width determine whether it is worth using our method by comparing the metrics of any dataset to those of the three involved datasets.

Utilizing the HiStruct information also for abstractive summarization is subject of future work.
Similarly, we see great potential in an encoder-decoder architecture with the proposed HiStruct injection components.

\section*{Acknowledgements}
The work presented in this article has received funding from the German Federal Ministry of Education and Research (BMBF) through the project QURATOR (no.~03WKDA1A).

\clearpage
\bibliography{my_acl}
\bibliographystyle{acl_natbib}
\clearpage

\appendix

\section{Appendix}
\label{sec:appendix}

\subsection{Statistics of the Datasets}
\label{subsec:Statistics of the datasets}

\begin{table}[ht]
\fontsize{9}{9}
\selectfont
\centering
\begin{tabular}{@{}lrrr@{}}
\toprule
\textbf{Dataset} &
  \textbf{CNN/DailyMail} &
  \textbf{PubMed} &
  \textbf{arXiv} \\ \midrule
 \multicolumn{4}{l}{Raw documents}                                                      \\ \midrule
 avg. \#words                  & 792.24          & 2,967.22 & 5,825.68  \\
 avg. \#sentences              & 40.31           & 86.37    & 206.3             \\
 avg. \#sections* & 31.2            & 5.91     & 5.55              \\
 avg.  hi-width                          & 1.33            & 15.79    & 37.33             \\ \midrule
 \multicolumn{4}{l}{Raw gold summaries}                                                           \\ \midrule 
avg. \#words              & 53.25           & 202.42   & 272               \\
avg. \#sentences          & 3.75            & 6.85     & 9.61              \\ \midrule 
\multicolumn{4}{l}{Novel n-grams in gold summaries}                                         \\ \midrule
avg. \% novel &&&\\
~~~~ 1grams                       & 13.97           & 0.2      & 0.15              \\
~~~~ 2grams                       & 51.79           & 2.69     & 2.73              \\ \midrule
\multicolumn{4}{l}{Nr. of documents} \\ \midrule 
\#train  & 287,227  & 119,924  & 203,037  \\
\#val    & 13,368   & 6,633    & 6,436  \\
\#test   & 11,490   & 6,658    & 6,440 \\\midrule
\multicolumn{4}{l}{Documents tokenized by the RoBERTa tokenizer} \\ \midrule
avg. doc length    & 964             & 4,252    & 8,991         \\
75\% doc length & 1,219           & 5,382    & 11,289            \\
85\% doc length & 1,448           & 6,709    & 14,294            \\
99\% doc length & 2,345           & 15,277   & 35,559            \\  \bottomrule
\end{tabular}
\caption[Statistics of the datasets]{Statistics of the datasets. * avg. \#paragraphs in CNN/DailyMail.}
\label{tab:datasets}
\end{table}

The CNN/DailyMail\footnote{\url{https://cs.nyu.edu/~kcho/DMQA/}}, PubMed and arXiv\footnote{\url{https://github.com/armancohan/long-summarization}} datasets are used in experiments. We use the original splits provided by \citet{cnndm} and \citet{pubmed} for training, validation and testing.

\subsection{Pre-defined Section Title Classes}
\label{subsec:Pre-defined section classes}

The pre-defined dictionaries of the typical section title classes and the corresponding in-class section titles are released in our GitHub project (see Section~\ref{sec:Introduction}). 
There are 164,195 unique section titles in PubMed, and 500,015 in arXiv, which are encoded as section title embeddings (STE) respectively using the proposed encoding method. 

For PubMed, we define 8 section title classes: introduction, background (i.\,e., background, review and related work), case (i.\,e., case reports), method, result, discussion, conclusion and additional information (i.\,e., additional information such as conflicts of interest, financial support and acknowledgments). For arXiv, we define 10 classes:  introduction, background, case, theory (i.\,e., problem formulation and proof of theorem), method, result, discussion, conclusion, reference and additional information. Classified STEs are prepared accordingly by replacing the original STEs of the intra-class section titles with the encoding of the section title class.

\subsection{Implementation Details}
\label{subsec:Implementation Details appendix}
\textbf{The learning rate schedule} follows \citet{presumm} with warm-up. On CNN/DailyMail, we train the HiStruct+ models and the TransformerETS baselines 50,000 steps with 10,000 warm-up steps. On PubMed and arXiv, the models are trained 70,000 steps with 10,000 warm-up steps when taking 15,000 tokens as input. When training models on arXiv with 28,000 input tokens, we train 100,000 steps with 10,000 warm-up steps. 

\textbf{The number of the extracted sentences} depends on the dataset. On CNN/DailyMail, we follow \citet{presumm} to select 3 sentences for each document as its extractive summary and apply Trigram Blocking \cite{paulus2018a} to reduce the redundancy of the selected sentences. On PubMed and arXiv, 7 sentences are extracted without Trigram Blocking.

\textbf{The length limit of the original TLM} is overcome by adding extra token linear position embeddings (tPE) to cover the desired length. The additional tPE are then trained with the whole summarization model. Instead of initializing them randomly, we copy the original tPE of the TLM multiple times until the desired length is covered. 

The HiStruct+ models and the TransformerETS baselines are trained on 3 GPUs (NVIDIA® Quadro RTX™ 6000 GPUs with 24GB memory) with gradient accumulation every two steps. Checkpoints are saved and evaluated on the validation set every 1,000 steps. The top-3 checkpoints based on the validation loss are kept. The batch size varies with the base TLM and the input length. On CNN/DailyMail, the base TLM is fine-tuned with the whole summarization model. Due to resource limitation, the TLM (i.\,e., Longformer) is not fine-tuned when training the summarization model on PubMed and arXiv with longer inputs. 

\subsection{Model Architectures  and Experimental Settings}
\label{subsec:Experimental settings}
The detailed model architectures and experimental settings for the models trained on CNN/DailyMail, PubMed and arXiv are summarized in Table \ref{tab:detailed_experimental_settings_on_cnndm_overview}, Table \ref{tab:detailed_experimental_settings_on_pubmed_overview} and Table \ref{tab:detailed_experimental_settings_on_arxiv_overview} respectively. 
The detailed model architectures and experimental settings include:
\begin{enumerate}[leftmargin=0pt,itemsep=0pt,parsep=0pt]
 \item[] \textbf{Base TLM}: the Transformer language model used for sentence encoding in the summarization system.
 \item[] \textbf{Input length}: how many tokens are taken as input.
 \item[] \textbf{Extra tPE}: how to initialize the extra input token linear position embeddings when taking longer input. We can choose to randomly initialize them or copy the original ones.
 \item[] \textbf{FT}: whether the base TLM is fine-tuned with the entire summarization model.
 \item[] \textbf{TL}: the number of the Transformer layers stacked upon the base TLM for extractive summarization.
 \item[] \textbf{WS}: warmup steps, how many steps are used for warming-up of the learning rate.
 \item[] \textbf{TS}: the total training steps.
 \item[] \textbf{BS}: batch size, how many documents are used as one batch during training.
 \item[] \textbf{AC}: accumulation count, gradient accumulation every $k$ steps.
 \item[] \textbf{GPU}: the number of GPUs used for training, we use NVIDIA® Quadro RTX™ 6000 GPUs with 24GB memory.
 \item[] \textbf{HiStruct}: the injection setting. Hierarchical structure information that can be injected into the summarization model are: sHE (i.\,e., sentence hierarchical position embeddings), STE (i.\,e., section title embeddings), or  STE(classified) (i.\,e., classified section title embeddings)
 \item[] \textbf{HPE}: the hierarchical position encoding method used in the model. The method is based on the sinusoidal (sin) or the learnable (la) linear position encoding method associated with a combination mode (i.\,e., sum/mean/concat)
 \item[] \textbf{\#PE}: the numbers of the learned position embeddings for each hierarchy-level of the hierarchical positions and the linear sentence positions, when using the learnable position encoding method. We set them to a same value during training.
 \item[] \textbf{SS}: saving steps, save checkpoints every  $k$ steps.
 \item[] \textbf{n}: select $n$ sentences as the extractive summary for each document.
 \item[] \textbf{TB}: trigram blocking, whether to apply Trigram Blocking during sentence selection
\end{enumerate}

\begin{table*}[ht]
\fontsize{9}{9}
\selectfont
\centering
\begin{tabular}{@{}llrcllll@{}}
\toprule
Models/Settings &
  Base TLM &
  \begin{tabular}[c]{@{}l@{}}Input \\ length\end{tabular} &
  \begin{tabular}[c]{@{}l@{}}Extra \\ tPE\end{tabular} &
  BS &
  HiStruct &
  HPE &
  \#PE 
  \\ \midrule
\multicolumn{8}{c}{Reproduced baselines}  \\ \midrule
TransformerETS &&& \\
~~ \textit{BERT-base (1,024 tok.)} & BERT-base    & 1,024 & copied & 200 &  none     & --       & --    \\
~~ \textit{BERT-large (512 tok.)}& BERT-large   & 512   & --      & 100 &  none     & --       & --    \\
~~ \textit{RoBERTa-base (1,024 tok.)}          & RoBERTa-base & 1,024 & copied & 250 &  none     & --       & --    \\ \midrule
\multicolumn{8}{c}{Our models (Extractive)} \\ \midrule
HiStruct+ &&& \\
~~ \textit{BERT-base (1,024 tok.)}     & BERT-base    & 1,024 & copied & 200 &  sHE only & la-sum  & 407  \\
~~ \textit{BERT-large (512 tok.)}     & BERT-large   & 512   & --      & 100 &  sHE only & la-sum  & 407  \\
~~ \textit{RoBERTa-base (1,024 tok.)} & RoBERTa-base & 1,024 & copied & 250 &  sHE only & la-sum  & 407  \\ \bottomrule
\end{tabular}
\caption[Experimental settings on CNN/DailyMail]{Detailed model architectures and experimental settings for models trained on CNN/DailyMail (also see Table \ref{tab:cnndm_result}). The settings not included in the table are the same for all models. FT: yes, TL:2, WS:10,000, TS:50,000, AC:2, GPU:3, SS:1,000, n: 3, TB:yes.}
\label{tab:detailed_experimental_settings_on_cnndm_overview}
\end{table*}

\begin{table*}
\fontsize{9}{9}
\selectfont
\centering
\begin{tabular}{@{}llrcll@{}}
\toprule
Models/Settings &
  Base TLM &
  BS &
  HiStruct &
  HPE &
  \#PE 
  \\ 
  \midrule
  
\multicolumn{6}{c}{Reproduced baselines} \\ \midrule
TransformerETS &&&&& \\
~~ \textit{Longformer-base (15k tok.)} &
  Longformer-base &
  500 &
  none &
  -- &
  -- \\
~~ \textit{Longformer-large (15k tok.)} &
  Longformer-large &
  256 &
  none &
  -- &
  -- \\ \midrule
\multicolumn{6}{c}{Our models (Extractive)} \\ \midrule
HiStruct+ &&&&& \\
~~ \textit{Longformer-base (15k tok.)} \\ 
~~~~~~ sHE+STE(classified) &
  Longformer-base &
  500 &
  sHE+STE(classified) &
  la-sum &
  450 \\
~~~~~~ sHE+STE &
  Longformer-base &
  500 &
  sHE+STE &
  la-sum &
  450\\
~~~~~~ sHE &
  Longformer-base &
  500 &
  sHE only &
  la-sum &
  450 \\ 
~~ \textit{Longformer-large (15k tok.)}\\ 
~~~~~~ sHE+STE(classified)&
  Longformer-large &
  256 &
  sHE+STE(classified) &
  la-sum &
  450 \\
~~~~~~ sHE &
  Longformer-large &
  256 &
  sHE only &
  la-sum &
  450 \\
 \bottomrule
\end{tabular}
\caption[Experimental settings on PubMed]{Detailed model architectures and experimental settings for models trained on PubMed (also see Table \ref{tab:pubmed_result}). The settings not included in the table are the same for all models. Input length: 15,000; Extra tPE: copied; FT: no; TL:2; WS:10,000; TS:70,000; AC:2; GPU:3; SS:1,000; n: 7; TB:no.}
  \label{tab:detailed_experimental_settings_on_pubmed_overview}
\end{table*}

\begin{table*}[ht]
\fontsize{9}{9}
\selectfont
\centering
\begin{tabular}{@{}llrrllrr@{}}
\toprule
Models/Settings &
Base TLM &
  \begin{tabular}[c]{@{}l@{}}Input \\ length\end{tabular} &
  TS &
  BS &
  HiStruct &
  HPE &
  \#PE 
\\ \midrule
\multicolumn{8}{c}{Reproduced baselines}\\ \midrule
TransformerETS &&&&& \\
~~ \textit{Longformer-base (15k tok.)}  & Longformer-base & 15,000  & 70,000  & 500  & none     & --      & --   \\
~~ \textit{Longformer-base (28k tok.)}  & Longformer-base & 28,000  & 100,000 & 500  & none     & --      & --  \\\midrule
\multicolumn{8}{c}{Our models (Extractive)}      \\ \midrule
HiStruct+ &&&&& \\                                          
~~ \textit{Longformer-base (15k tok.)} \\
~~~~~~ sHE+STE(classified)         & Longformer-base  & 15,000  & 70,000  & 500  & sHE+STE(classified) & la-sum & 720  \\
~~~~~~ sHE+STE & Longformer-base  & 15,000  & 70,000  & 500  & sHE+STE             & la-sum & 720  \\
~~~~~~ sHE   & Longformer-base           & 15,000 & 70,000  & 500  & sHE only            & la-sum & 720  \\ 
~~ \textit{Longformer-base (28k tok.)} \\                                                           
~~~~~~ sHE+STE(classified)  & Longformer-base     & 28,000 & 100,000 & 500  & sHE+STE(classified) & la-sum & 1300  \\
~~~~~~ sHE+STE & Longformer-base   & 28,000 & 100,000 & 500  & sHE+STE             & la-sum & 1300 \\ \bottomrule
\end{tabular}
\caption[Experimental settings on arXiv]{Detailed model architectures and experimental settings for models trained on arXiv (also see Table \ref{tab:arxiv_result}). The settings not included in the table are the same for all models. Extra tPE: copied; FT: no; TL:2; WS:10,000; AC:2; GPU:3; SS:1,000; n: 7; TB:no.}
  \label{tab:detailed_experimental_settings_on_arxiv_overview}
\end{table*}

\subsection{Ablation Studies on CNN/DailyMail}
\label{subsec:Ablation studies on CNN/DailyMail}

\begin{table}[ht]
\fontsize{9}{9}
\selectfont
\centering
\begin{tabular}{@{}llll@{}}
\toprule
Experimental Results  &  \multicolumn{1}{c}{R1} & \multicolumn{1}{c}{R2} & \multicolumn{1}{c}{RL}    \\ \midrule
BERT-base (512 tok.)       \\ \midrule
HiStruct(sHE)+      & \underline{43.23} & \underline{20.15} & \underline{39.65} \\
HiStruct(sHE\&tHE)+ & 40.76 & 18.03 & 37.08 \\ \midrule
BERT-base (1,024 tok.)  & & &        \\ \midrule
HiStruct(sHE)+      & \underline{43.38} & \underline{20.33} & \underline{39.78} \\
HiStruct(sHE\&tHE)+ & 41.04 & 18.25 & 37.41 \\ \midrule
BERT-large (512 tok.) & & &         \\ \midrule
HiStruct(sHE)+      & \underline{43.46} & \underline{20.4}  & \underline{39.85} \\
HiStruct(sHE\&tHE)+ & 40.58 & 17.71 & 36.83 \\ \bottomrule
\end{tabular}
\caption[Ablation study on CNN/DailyMail (a)]{Ablation study on CNN/DailyMail (a). Comparison of HiStruct+ models with/without token-level hierarchical position embeddings (tHE). The models in different blocks are based on different TLMs with various input lengths. Underlined are the best ROUGEs in each block.}
\label{tab:cnndm_ablation_a}
\end{table}

\textbf{The effect of token-level hierarchical position embeddings} is investigated in experiments. The hierarchical position embeddings of tokens are generated as followings: 

Given the $t$-th token within the document, its hierarchical position is represented by Equation \ref{eq:tokstructurevec}:
\begin{equation} \label{eq:tokstructurevec}
TSV_t = (a_t, b_t, c_t)
\end{equation}
where $a_t$ represents the linear position of the section  which contains the token, $b_t$ is the sentence's position within the section and $c_t$ is the linear position of the token within the sentence. 

Given the $t$-th token and the desired size of the output embeddings $d$, its token hierarchical position embeddings (tHE) is encoded by Equations \ref{eq:tHE_sum}, \ref{eq:tHE_mean}, \ref{eq:tHE_concat}, using different combination modes.
\begin{equation} \label{eq:tHE_sum}
\footnotesize
tHE_{\text{sum}} (t, d) =PE(a_t,d)+PE(b_t,d)+PE(c_t,d)
\end{equation}
\begin{equation}\label{eq:tHE_mean}
\footnotesize
tHE_{\text{mean}} (t, d) =\frac{PE(a_t,d)+PE(b_t,d)+PE(c_t,d)}{3}
\end{equation}
\begin{equation}\label{eq:tHE_concat}
\footnotesize
tHE_{\text{concat}} (t, d) =PE(a_t,\frac{d}{3}) | PE(b_t,\frac{d}{3}) | PE(c_t,\frac{d}{3})
\end{equation}

Initial experiments are conducted to assess the summarization performance of the HiStruct+ models with or without the tHE. For this purpose, we compare a HiStruct+ model merely injecting sentence hierarchical position embeddings (i.\,e., sHE) with a HiStruct+ model with both sentence and token hierarchical position embeddings (i.\,e., sHE\& tHE). That is, it adds the corresponding tHEs to the input embeddings at each input position, which are fed into the TLM. It also injects sHEs into the output sentence representations.

Table \ref{tab:cnndm_ablation_a} summarizes the evaluation results of three groups of HiStruct+ models based on different TLM with various input lengths. In each group, all experimental settings and parameters are the same, except for the injection setting of tHE.
The experimental results suggest that the HiStruct+ models with merely sHE consistently outperform the HiStruct+ models with both sHE \& tHE under various circumstances. The reason might be that we directly fine-tune the TLM on the extractive summarization task. When adding extra tHE to the input embeddings to the TLM, we do not pre-train the TLM with the adjusted inputs. It is reasonable that the TLM has difficulties in understanding of the new inputs based on the knowledge learned from the original format of encoding. Previous works, such as LayoutLM \cite{layoutlm}, LamBERT \cite{lambert} and HIBERT \cite{zhang-etal-2019-hibert},  which adjust the input embeddings or the encoder architecture of the pre-trained TLM, continue to pre-train the TLM on their own data. Continuing pre-training of the language models is a core part of these works and leads to significant improvements on downstream tasks. Due to lack of computing resources, we are not able to pre-train the language models. Furthermore, the key goal of our work is to experiment with various methods to make use of the internal hierarchical text structure information for extractive summarization. In this work, we conduct further experiments without token-level hierarchical position information and leave for future work the pre-training of language models with the adjusted input embeddings.

\begin{table}[ht]
\fontsize{9}{9}
\selectfont
\centering
\begin{tabular}{@{}lllllll@{}}
\toprule
\multirow{2}{*}{} &
  \multicolumn{3}{c}{la PE} &
  \multicolumn{3}{c}{sin PE} \\ \cmidrule(l){2-7} &
  \multicolumn{1}{c}{R1} &
  \multicolumn{1}{c}{R2} &
  \multicolumn{1}{c}{RL} &
  \multicolumn{1}{c}{R1} &
  \multicolumn{1}{c}{R2} &
  \multicolumn{1}{c}{RL} \\ \midrule
\multicolumn{7}{c}{HiStruct+BERT-base (1,024 tok.)}   \\ \midrule
sum    & \underline{43.38} & \underline{20.33} & \underline{39.78} & \underline{43.37} & 20.27 & \underline{39.75} \\
mean   & 43.33 & 20.31 & 39.73 & 43.33 & 20.28 & 39.72 \\
concat & 43.22 & 20.18 & 39.61 & \underline{43.37} & \underline{20.29} & 39.74 \\ \bottomrule
\end{tabular}
\caption[Ablation study on CNN/DailyMail (b)]{Ablation study on CNN/DailyMail (b). Comparison of HiStruct+ models using various hierarchical position encoding methods based on the sinusoidal or the learnable PE method, associated with the combination modes of sum, mean and concat respectively. Underlined are the best ROUGEs in each block.} %
\label{tab:cnndm_ablation_b}
\end{table}

\textbf{The effect of different settings for hierarchical position encoding} is investigated. As explained previously, based on different position encoding (PE) methods  (i.\,e., sin or la) associated with various combination modes (i.\,e., sum, mean or concat), we have totally six different settings for hierarchical position encoding: sin-sum, sin-mean, sin-concat, la-sum, la-mean and la-concat.  We investigate the effect of those 6 encoding settings systematically in experiments while keeping the rest settings and parameters the same, so that the evaluation results are comparable.

Table \ref{tab:cnndm_ablation_b} summarizes the evaluation results of six HiStruct+ models using the six encoding settings respectively, which are all trained on CNN/DailyMail based on BERT-base with 1,024 input tokens, injecting merely sHE. %
We observe that when using the la method, the combination mode sum leads to better results compared to the other modes (see the first three columns in Table \ref{tab:cnndm_ablation_b}). When using the sin method,  the various combination modes do not make a conspicuous difference in summarization performance. The sum and concat modes perform slightly better. When using the sum mode, the la and the sin methods produce similar results (see the first row in Table \ref{tab:cnndm_ablation_b}).

\begin{table}[ht]
\fontsize{9}{9}
\selectfont
\centering
\begin{tabular}{@{}llll@{}}
\toprule
Experimental Results & \multicolumn{1}{c}{R1} & \multicolumn{1}{c}{R2} & \multicolumn{1}{c}{RL}  \\ \midrule
BERT-base (1,024 tok.) & & &        \\ \midrule
TransformerETS            & 43.32 & 20.27 & 39.69 \\
HiStruct(la-sum)+  & \underline{43.38} & \underline{20.33} & \underline{39.78} \\
HiStruct(sin-sum)+ & 43.37 & 20.27 & 39.75 \\ \midrule
BERT-large (512 tok.) & & &          \\ \midrule
TransformerETS             & 43.45 & 20.36 & 39.83 \\
HiStruct(la-sum)+  & \underline{43.49} & \underline{20.4}  & \underline{39.9}  \\
HiStruct(sin-sum)+ & 43.46 & \underline{20.4}  & 39.85 \\ \midrule
RoBERTa-base (1,024 tok.) & & &      \\ \midrule
TransformerETS            & 43.62 & 20.53 & 39.99 \\
HiStruct(la-sum)+  & \underline{43.65} & 20.54 & \underline{40.03} \\
HiStruct(sin-sum)+ & 43.64 & \underline{20.56} & 40.02 \\ \bottomrule
\end{tabular}
\caption[Ablation study on CNN/DailyMail (c)]{Ablation study on CNN/DailyMail (c). Comparison of the TransformerETS baseline and the HiStruct+ models using the la-sum and sin-sum settings for hierarchical position encoding respectively. The models in different blocks are based on different TLMs with various input lengths. Underlined are the best ROUGEs in each block.}
\label{tab:cnndm_ablation_c}
\end{table}

\textbf{The effect of the encoding settings la-sum  vs. sin-sum} is further investigated in experiments. As discussed above, the encoding settings la-sum and sin-sum lead to similar results. We conduct experiments to further investigate the effect of using these methods. We also compare our HiStruct+ models with the corresponding TransformerETS baseline which differs from our models only in that it does not take into account extra HiStruct information. 

Table \ref{tab:cnndm_ablation_c} includes the ROUGEs of three set of comparison models, which use different TLM with various input lengths. In each group, the first model is the baseline without HiStruct injection. The second model and the third model differ from each other only with regard to the encoding setting. The experimental results suggest that both of the settings improve the summarization performance of the baseline model.
It is also observed that the la-sum method outperforms the sin-sum method slightly on CNN/DailyMail. The differences are not substantial.

\begin{table}[ht!]
\fontsize{9}{9}
\selectfont
\centering
\begin{tabular}{@{}llll@{}}
\toprule
Experimental Results & \multicolumn{1}{c}{R1} & \multicolumn{1}{c}{R2} & \multicolumn{1}{c}{RL} \\ \midrule
HiStruct(sHE)+ &&&\\
BERT-base (1,024 tok.) & & &  \\ \midrule
-\#Transformer layers \\
for summarization & & &         \\ \midrule
1                           & 43.29       & 20.25      & 39.69      \\
2                           & \underline{43.37}       & \underline{20.27}      & \underline{39.75}      \\
3                           & 43.16       & 20.15      & 39.56      \\ \midrule
-Extra Token Linear \\ Position Embeddings (tPE) & & &     \\ \midrule
Randomly initialized        & 40.53       & 17.76      & 36.8       \\
Copied                      & \underline{43.37}       & \underline{20.27}      & \underline{39.75}  \\ \midrule
-With/without Sentence Linear \\ Position Embeddings (sPE) & & &  \\ \midrule
With sPE                    & \underline{43.37}       & \underline{20.27}      & \underline{39.75}      \\
Without sPE                 & 43.31       & 20.25      & 39.69      \\ \bottomrule
\end{tabular}
\caption[Ablation study on CNN/DailyMail (d)]{Ablation study on CNN/DailyMail (d). The first block deals with the variation of the numbers of inter-sentence Transformer layers stacked on top of the TLM. The second block deals with the different methods to initialize extra input Token Linear Position Embeddings when taking longer input. The third block deals with the effect of Sentence Linear Position Embeddings. Underlined are the best ROUGEs in each block. }
\label{tab:cnndm_ablation_d}
\end{table}

\textbf{The effect of the number of the stacked Transformer layers} is investigated in our experiments. 
We fine-tune an extended BERT-base model with 1,024 input tokens for extractive summarization. %
We construct the HiStruct+ models with 1, 2, 3 stacked Transformer layers respectively, while keeping all other settings the same.  The results of those three HiStruct+ models are reported in the first block in Table \ref{tab:cnndm_ablation_d}. %
It is suggested that two stacked Transformer layers perform best in our HiStruct+ models for extractive summarization.

\textbf{The effect of different initialization strategies for the additional input Token Linear Position Embeddings} is also investigated in experiments. When taking input texts longer than the original input length of the base TLM, we need to add extra Token Linear Position Embeddings (tPE) for each extended position. We can choose to randomly initialize the extra tPE or copy the original ones to cover the extended input length. To investigate the effect of different initialization strategies, we use the basic settings of the HiStruct+ model with two summarization layers, namely the second model in the first block in Table \ref{tab:cnndm_ablation_d}. To build the comparison model, only the initialization strategy is changed to random.  As shown in the second block in Table \ref{tab:cnndm_ablation_d}, substantial improvements are achieved by using the copied tPEs for initialization instead of random initialization. ROUGE-1, ROUGE-2 and ROUGE-L are increased by 2.84, 2.51 and 2.95 respectively. 

\textbf{The effect of the Sentence Linear Position Embeddings} is also investigated in experiments. As shown in Figure \ref{fig:overview}, besides the hierarchical positions of each sentence, we also take the linear position of each sentence within the whole document into account by adding a Sentence Linear Position Embedding (sPE) to each sentence representation.  We assess the effect of the sPE by comparing two HiStruct+ models with or without the sPE. The second model in the first block in Table \ref{tab:cnndm_ablation_d} is compared to a model that differs from it only in the injection of sPE. The results are shown in the third block in Table \ref{tab:cnndm_ablation_d}. The HiStruct+ model with sPE outperforms the HiStruct+ model without sPE by a small margin regarding all ROUGE metrics.

\subsection{Human Evaluation of Extracted Summaries}
\label{subsec:human_evluation}
\begin{figure*}[ht]
\begin{subfigure}{1\textwidth}
  \centering
  \includegraphics[width=1\textwidth]{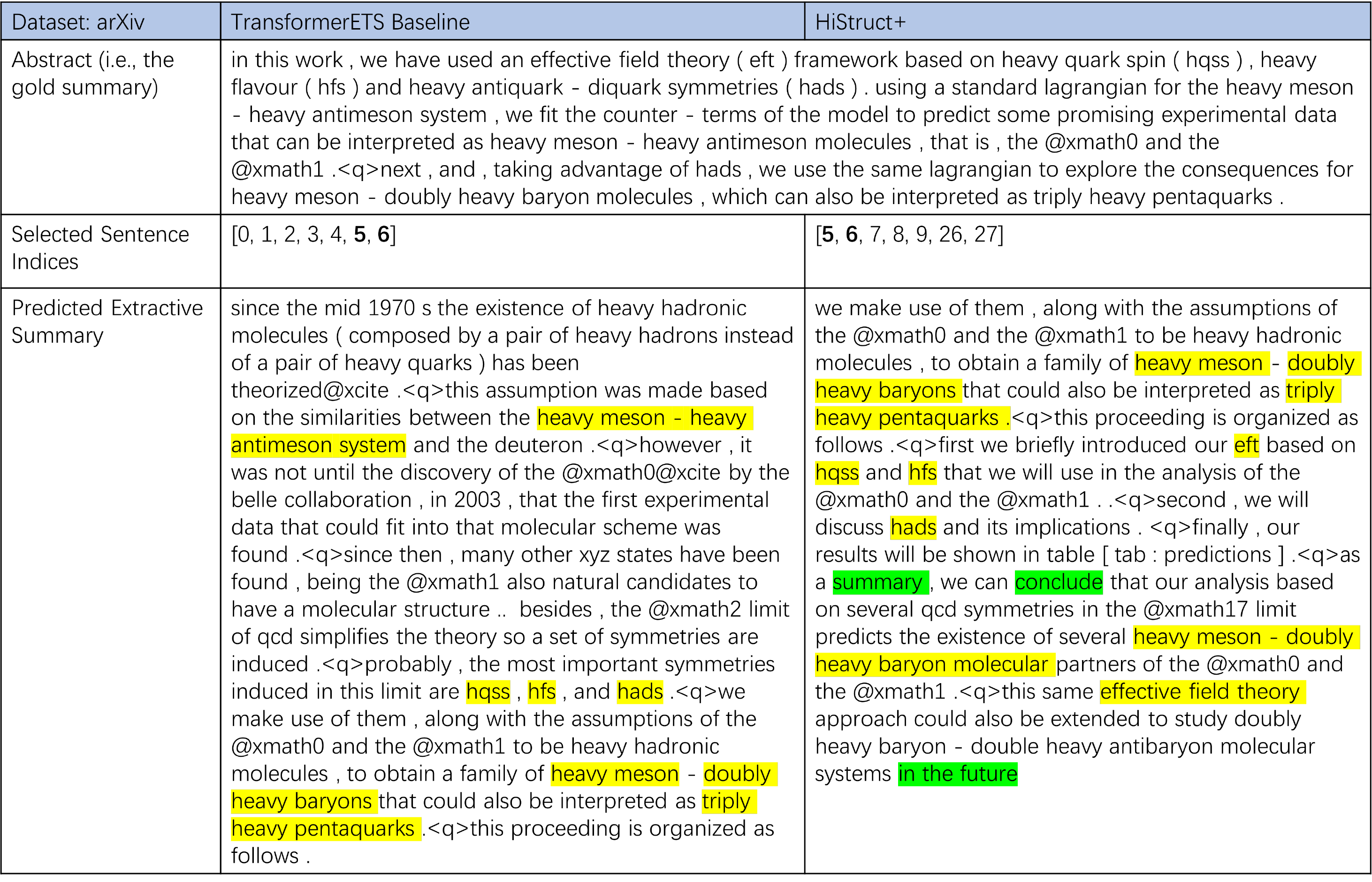}
  \label{fig:human_eval1.1}
\end{subfigure}
\begin{subfigure}{1\textwidth}
  \centering
  \includegraphics[width=1\textwidth]{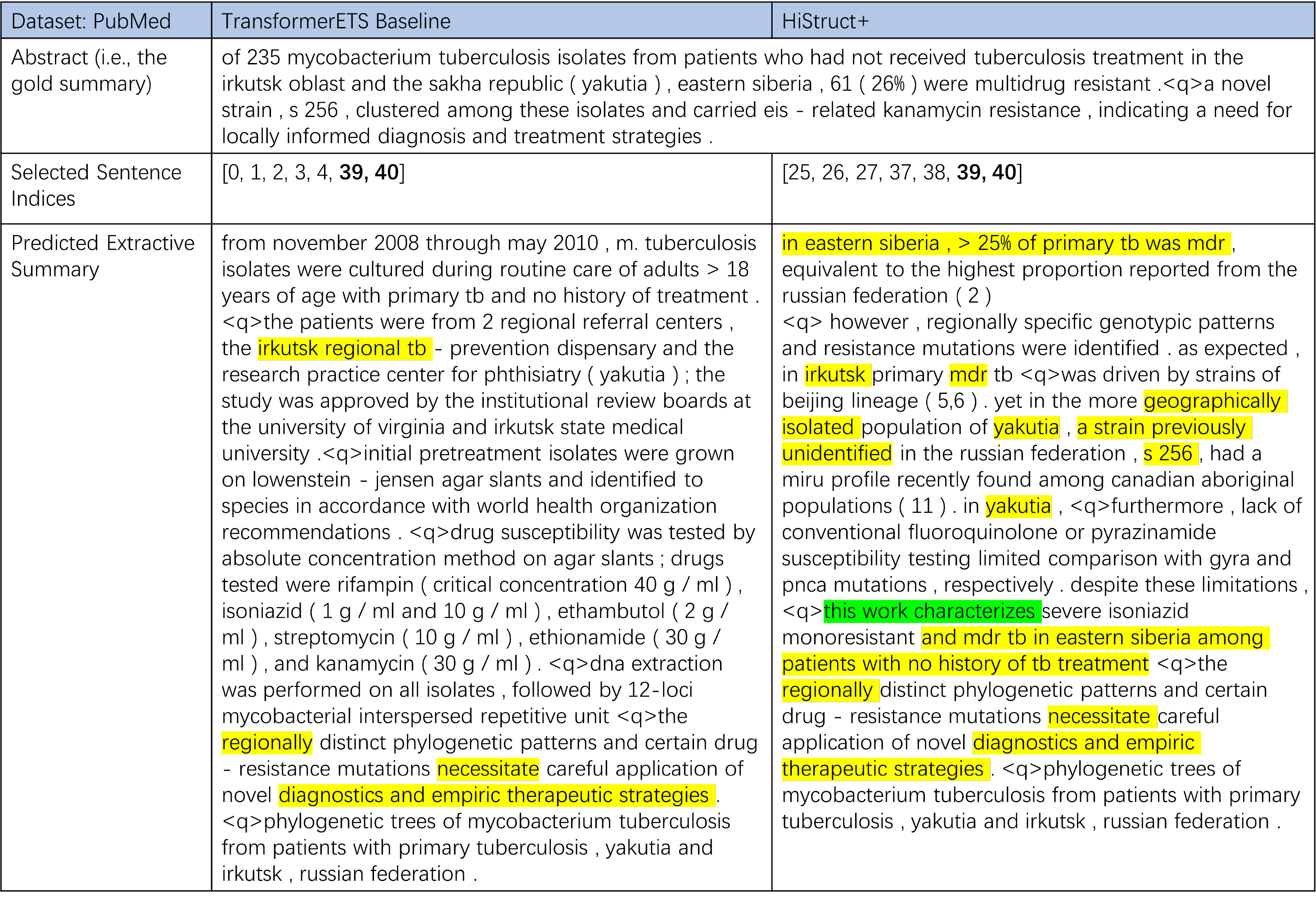}
  \label{fig:human_eval2}
\end{subfigure}
\caption{Two samples for human evaluation and case analysis of the extractive summaries predicted by the HiStruct+ model and the baseline model, in comparison with the gold summary (i.e., the abstract of the paper). The first sample is selected from the arXiv dataset, while the second sample is from PubMed. Top-7 sentences with the highest predicted scores are extracted, and then combined in their original order to construct a final summary. Their linear indices within the original document are shown in the second row of each table. The texts highlighted in yellow are the key words and the main content that appear in the gold summary. The phrases highlighted in green indicate typical parts of a scientific paper such as summary and future work. Sentences are split by '<q>'.}
\label{fig:human evaluation}
\end{figure*}

To have a more intuitive understanding about the superiority of the proposed system, we showcase two samples in Figure \ref{fig:human evaluation} for human evaluation and case analysis. The extractive summaries predicted by the HiStruct+ model and the baseline model are demonstrated respectively, in comparison with the gold summary (i.e., the abstract of the paper). To construct a final summary, top-7 sentences with the highest scores predicted by the model are extracted, and then combined in their original order.

The first arXiv sample shows that the baseline simply selects the first sentences. The predicted summary focuses on detailed background knowledge and lacks an overview of the proposed work. In contrast, our HiStruct+ model selects sentences at later positions. The first five sentences introduce the main content from an overall perspective. The last two sentences draw conclusions and give an outlook to future work, which is indicated by the phrases highlighted in green. 

The PubMed sample also indicates that the baseline favors the first sentences, which is consistent with our observations in Figure \ref{fig:extracted summaries}. Although the last two sentences highlight the same conclusion as in the gold summary that locally informed diagnosis and treatment strategies are needed, too much background information is unnecessarily included in the first five sentences. Our HiStruct+ model selects more informative sentences at later positions. The predicted summary covers all key parts of the gold summary: 1). the statistics are reported (i.e.,  26\% of primary tuberculosis (tb) was multidrug resistant (mdr)); 2). the novel strain s256 is mentioned; 3). the conclusion is highlighted. The overall topic of the work is especially highlighted by the sentence with the green-colored phrase.

\clearpage

\end{document}